\documentclass[letterpaper, 10 pt, conference]{styles/ieeeconf}  

\IEEEoverridecommandlockouts                              

\usepackage{times}
\usepackage{blindtext, rotating}

\usepackage{epsfig}
\usepackage{graphicx}
\usepackage{amsmath}
\usepackage{amssymb}

\usepackage{tabularx}

\usepackage[utf8]{inputenc}
\usepackage[T1]{fontenc}
\usepackage{textcomp}
\usepackage{gensymb}
\usepackage{multirow}
\usepackage{booktabs}
\usepackage{xcolor}
\usepackage{graphicx}
\usepackage{subcaption}
\usepackage{float}
\usepackage{capt-of}
\usepackage{etoolbox}
\usepackage{adjustbox}
\usepackage[font=small]{caption}

\usepackage{paralist}

\usepackage[colorlinks=true, pagebackref]{hyperref}

\usepackage{epigraph}

\newcommand{\etal}{\emph{et al.}}
\newcommand{\cf}{\emph{cf.}}

\newcommand{\monolayout}{\emph{MonoLayout}}
\newcommand{\videolayout}{\emph{VideoLayout}}
\newcommand{\bev}{bird's eye view}
\newcommand{\coolname}{\emph{AutoLay}}
\newcommand{\autolay}{\emph{AutoLay}}

\newcommand{\mb}[1]{\mathbf{#1}}
\newcommand{\mcal}[1]{\mathcal{#1}}

\title{\textbf{AutoLay: Benchmarking amodal layout estimation for autonomous driving}}

\usepackage{authblk}

\makeatletter
\renewcommand\AB@affilsepx{, \protect\Affilfont}
\makeatother

\author[1,2]{Kaustubh Mani$^*$\thanks{$^*$ denotes equal contribution}\thanks{Corresponding author: \url{kaustubh3095@gmail.com} }}
\author[1]{N. Sai Shankar$^*$}
\author[3,4]{Krishna Murthy Jatavallabhula}
\author[1,2]{K. Madhava Krishna}

\affil[1]{\small{Robotics Research Center, KCIS}}
\affil[2]{\small{IIIT Hyderabad}}
\affil[3]{\small{Mila - Quebec AI Institute, Montreal}}
\affil[4]{\small{Universit\'e de Montr\'eal}}

\begin{document}
\bstctlcite{Force_Etal}

\makeatletter
\let\@oldmaketitle\@maketitle
\renewcommand{\@maketitle}{\@oldmaketitle
\centering
\includegraphics[scale=0.3]{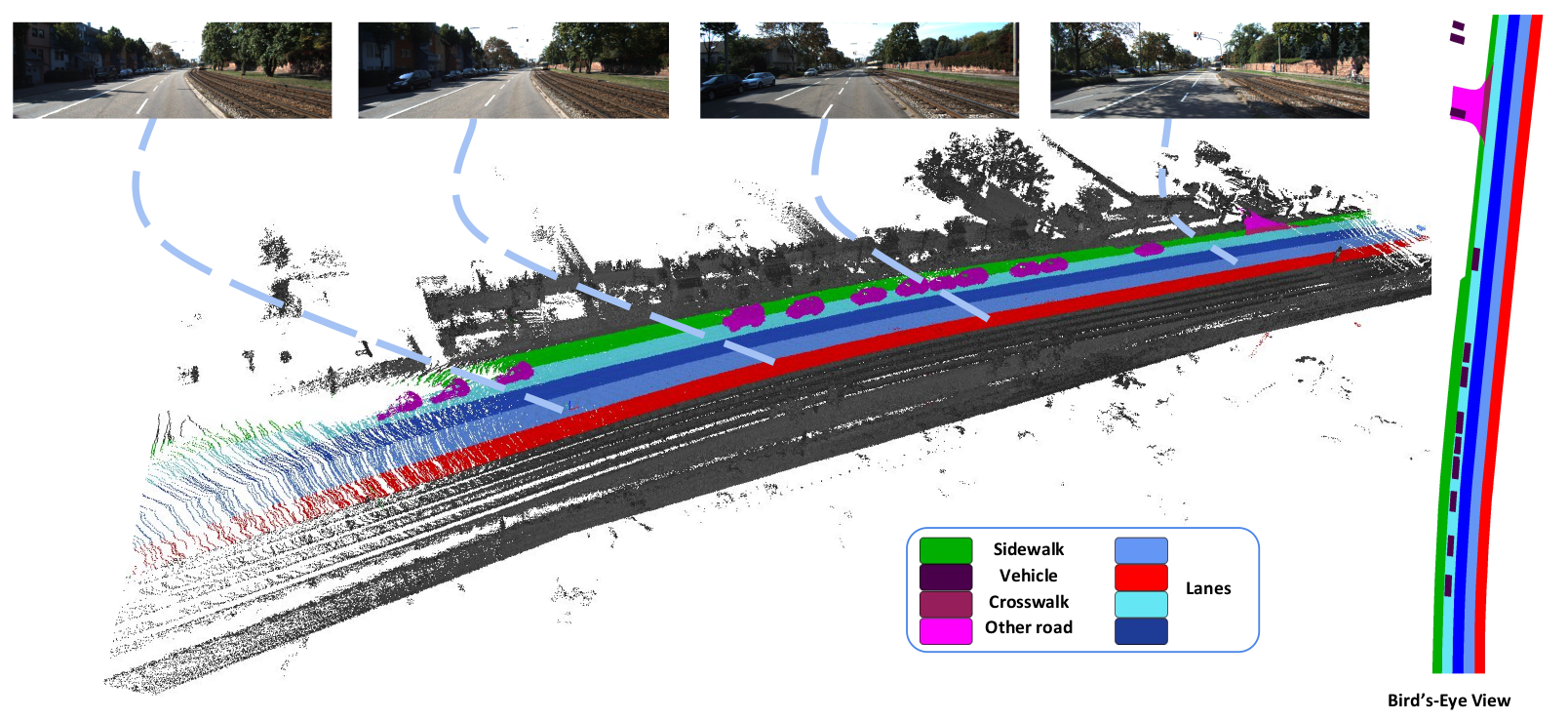}

\captionof{figure}{ 
\textbf{Amodal layout estimation} is the task of estimating a semantic occupancy map in \bev{}, given a monocular image or video. The term \emph{amodal} implies that we estimate occupancy and semantic labels even for parts of the world that are occluded in image space. In this work, we introduce \textbf{\coolname{}}, a new dataset and benchmark for this task. \coolname{} provides annotations in 3D, in \bev{}, and in image space. A sample annotated sequence (from the KITTI dataset~\cite{KITTI}) is shown below. We provide high quality labels for sidewalks, vehicles, crosswalks, and lanes. We evaluate several approaches on sequences from the KITTI~\cite{KITTI} and Argoverse~\cite{chang2019argoverse} datasets.
}
\label{fig:teaser_new}
\vspace{0.2cm}
}

\makeatother

\maketitle
\begin{abstract}

Given an image or a video captured from a monocular camera, amodal layout estimation is the task of predicting semantics and occupancy in \bev{}. The term \emph{amodal} implies we also reason about entities in the scene that are occluded or truncated in image space.
While several recent efforts have tackled this problem, there is a lack of standardization in task specification, datasets, and evaluation protocols.
We address these gaps with \coolname{}, a dataset and benchmark for amodal layout estimation from monocular images.
\coolname{} encompasses driving imagery from two popular datasets: KITTI~\cite{KITTI} and Argoverse~\cite{chang2019argoverse}.
In addition to fine-grained attributes such as lanes, sidewalks, and vehicles, we also provide semantically annotated 3D point clouds. We implement several baselines and bleeding edge approaches, and release our data and code.\footnote{Project page: \url{https://hbutsuak95.github.io/AutoLay/}}.

\end{abstract}

\setcounter{figure}{1} 
\section{Introduction}
\label{sec:introduction}

    
Commercial interest in autonomous driving has led to the emergence of several interesting and challenging problems, particularly in terms of perception. In this work, we focus on the problem of \emph{amodal scene layout estimation} (introduced in~\cite{monolayout}). \emph{Amodal} perception, as studied by cognitive scientists, refers to the phenomenon of ``imagining" or ``hallucinating" parts of the scene that do not result in sensory stimulation. In the context of urban driving, we formulate the \emph{amodal layout estimation} task as predicting a \bev{} semantic map from monocular images that explicitly reasons about entities in the scene that are not visible from the image(s).

Amodal scene layout estimation is an emerging area of study where prior efforts~\cite{monolayout, monooccupancy, schulter2018learning} have only focused on amodal completion for \emph{coarse} categories such as roads and sidewalks. While recent efforts have incorporated dynamic objects~\cite{monolayout}, they fall short of modeling higher-level behaviors in urban driving scenarios such as lane-keeping, merging, etc. Further, each of these efforts~\cite{monolayout, monooccupancy, schulter2018learning} lack consensus in problem specifications and evaluation protocols.

We address both the above issues. We first formalize the problem of \emph{amodal scene layout estimation} that unifies existing work and extends the task to several fine-grained categories. To foster research in this area, we provide \coolname{}, a dataset comprising of over $16000$ images over a distance of $12$ kilometers, annotated with fine-grained \emph{amodal} completions in 3D as well as in \bev. \coolname{} includes $44$ sequences from KITTI~\cite{KITTI} and $89$ sequences from the Argoverse~\cite{chang2019argoverse} datasets. We implement several baselines and prior art, and open-source our code, models, and data~\cite{monolayout, monooccupancy, you2019pseudo}. We additionally propose \videolayout{}, a simple approach that achieves top performance on the \coolname{} benchmark by leveraging temporal information across image sequences, advancing the state-of-the-art of amodal layout estimation.




\begin{figure*}[!ht]
  \centering
  \includegraphics[width=\textwidth]{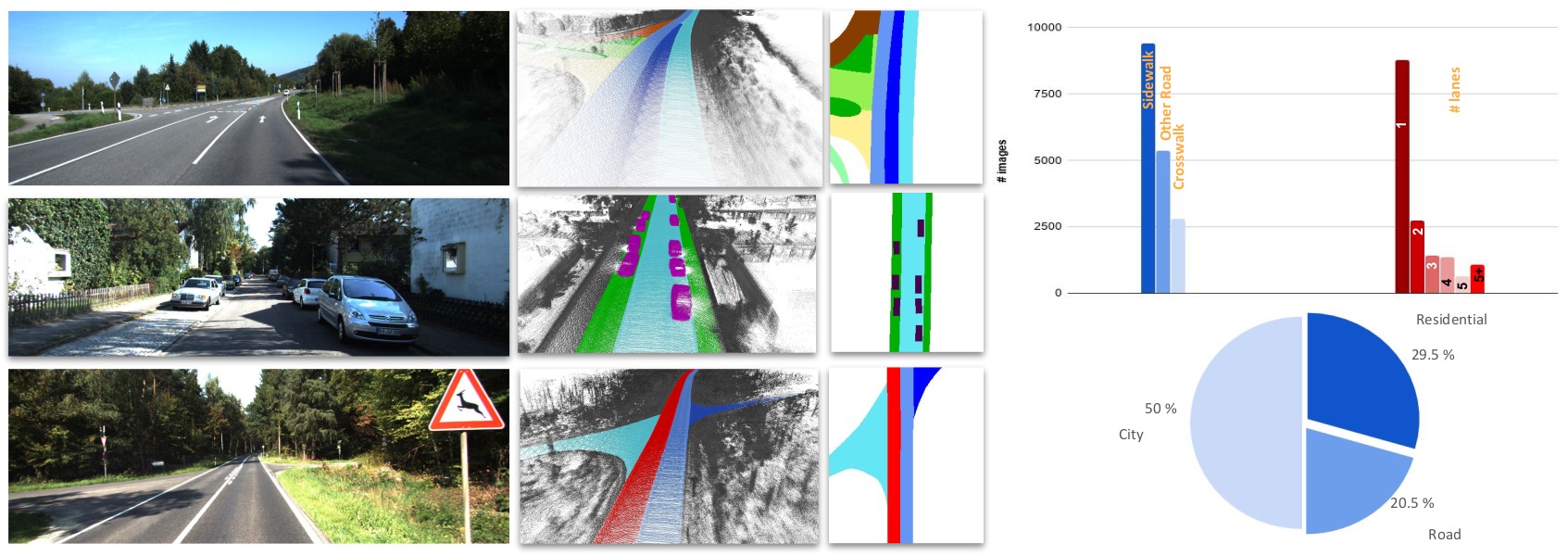}
  \captionof{figure}{\textbf{Dataset}: (\emph{Left to right}) Sample images from the \emph{KITTI} split of \coolname{}. Corresponding annotated lidar pointclouds. Amodal scene layout in \bev{}. (\emph{Last column}) Distribution of semantic classes (bar plot), and scene types (pie chart).
}
  \label{fig:dataset}
\end{figure*}



    
    
    
    


\section{Related work}
\label{sec:relatedwork}


Interpreting 3D scenes from an RGB image has been a longstanding challenge for computer vision systems. In this section, we enlist a few approaches to \emph{amodal} perception. For a comprehensive survey of general 3D scene understanding, we refer the interested reader to Ozyecsil \etal~\cite{ozyecsil2017survey}.

\subsubsection*{\textbf{Amodal perception for indoor scene understanding}}

In the context of indoor 3D scene understanding, a few approaches to \emph{amodal} perception have been proposed~\cite{amodalKTCM15, factored3dTulsiani17}. While these approaches use coarse voxel grids to represent objects, we instead focus on an orthographic top-down view that restricts semantic understanding to road regions. We note that most of the semantic understanding necessary for driving-related tasks can be obtained on the road regions, making this simplification attractive.

\subsubsection*{\textbf{Amodal perception for autonomous driving}}

To the best of our knowledge, there are no previous works that provide a unified approach to obtain an amodal scene layout estimation in \bev{} for fine-grained static and dynamic classes like lanes, sidewalks, vehicles, etc. Existing approaches~\cite{monooccupancy, schulter2018learning, wang2019parametric, lu2019hallucinating} reason about ``coarse-grained" static classes such as roads and sidewalks. Schulter \etal~\cite{schulter2018learning} obtain occlusion-aware \bev{} road layouts by hallucinating behind objects tagged as \emph{foreground} in an image.
Wang \etal~\cite{wang2019parametric} develop a parametric representation of road scenes that reason about the number of lanes, width of each lane, presence of an intersection, etc.
Lu \etal~\cite{monooccupancy} use a variational autoencoder to predict road, sidewalk, terrain, and non free-space in bird's eye view.
Lu~\etal{}~\cite{lu2019hallucinating} also perform vehicle shape completion, albeit independent of the static layout estimation task.
Garnett~\etal{}~\cite{garnett20193d} predict static road and lane layouts by estimating lane boundaries in 3D.

For dynamic scene understanding, several approaches aim to detect object layouts in 3D. While some of these~\cite{ku2018joint, chen2017multi, liang2018deep} combine information from images and lidar, others~\cite{yang2018pixor, roddick2018orthographic, wang2019pseudo} operate by converting images to \bev{} representations, followed by object detection. However, these techniques are often developed independently of static layout estimation methods.

In earlier work, we presented MonoLayout~\cite{monolayout}- perhaps the first approach to amodally reason about the static and dynamic layout of an urban driving scene. While the interest in this nascent area is rapidly increasing, the absence of standardized task specification and evaluation have been slowing down research. This forms the primary motivation behind the \coolname{} dataset and benchmark.

 


\section{Amodal scene layout estimation}
\label{sec:problem}

We begin by formally defining the task of amodal scene layout estimation. Given an image (or a sequence of images) $\mcal{I}$, usually in perspective view, we wish to learn a \emph{labeling function} $\Phi$ of \emph{all} points within an area $\mcal{R}$ (usually rectangular) around the camera. The region of interest $\mcal{R}$ is often in a different view than that of the perspective image. In this paper, we focus on \bev{}, which refers to a top-down orthographic view of the ground plane. The labeling function $\Phi$ must produce a label distribution (over a set of $N$ classes) for world points within the swath $\mcal{R}$, \emph{regardless of whether or not they are imaged in} $\mcal{I}$.

In particular, we propose the estimation of the following semantic classes tasks in the context of this problem: road, sidewalk, crosswalk, lanes (ego-lane, other lanes), other road, vehicle. Depending on the exact setup used (single-image vs image-sequences, semantic categories to be estimated, etc.), this enables fair evaluation of all approaches. We defer a detailed description of evaluation protocols and its challenges to Sec.~\ref{sec:experiments}.




\section{AutoLay Dataset}

Over the last decade, several public datasets~\cite{KITTI, oxfordrobotcar, nuscenes2019, apolloscape, chang2019argoverse} have enabled massive strides in autonomous driving research. While nearly all the above datasets provide out-of-the-box support (annotations, evaluation utilities, metrics, benchmarking support) for popular tasks like object detection, semantic segmentation, multi-object tracking, trajectory forecasting, none of these currently benchmark amodal scene layout estimation.

Obtaining training data for the task of amodal layout estimation involves multiple challenges. It involves perceiving scene segments that are not even imaged by the camera. 
While most datasets have lidar scans that often span a larger sensor swath, such lidar beams are not dense enough to perceive thin scene structures.
Approaches such as \monolayout{}~\cite{monolayout} compensate for this lack of precision training data by adversarial learning, using OpenStreetMap~\cite{OpenStreetMap} data.
More recently, Wang \etal~\cite{wang2019parametric} release a dataset that provides a \emph{parametric} \bev{} representation for images from the KITTI~\cite{KITTI} and nuScenes~\cite{nuscenes2019} datasets, but the top-view representations are synthesized views of a simulated parametric surface model. In summary, there is currently no dataset that provides off-the-shelf support for benchmarking amodal layout estimation. We fill this void with \coolname, a dataset for amodal layout estimation in \bev{} (Fig.~\ref{fig:teaser_new}).

\subsection{Dataset overview}

We use $44$ video sequences from the KITTI Raw dataset~\cite{KITTI} in \coolname. We provide per-frame annotations in perspective, orthographic (\bev), as well as in 3D. Of the $44$ annotated sequences, $24$ sequences---containing $10705$ images---are used for training. The other $20$ sequences---comprising $5626$ images---form the test set. This makes for nearly $16$K annotated images, across a distance of $12$ Km, and a variety of urban scenarios (\emph{residential}, \emph{city}, \emph{road}) (\cf~Fig~\ref{fig:dataset}).
The semantic classes considered in this dataset are \emph{road}, \emph{sidewalk}, \emph{vehicle}, \emph{crosswalk}, and \emph{lane}.
Each \emph{lane} segment is provided a unique id, which we classify further\footnote{We perform this subclassification for our specific task, by projecting lane segments to \bev.}
The \emph{lane} class is further classified as \emph{ego-lane} and \emph{other lane}. We also have an \emph{other road} class for road areas that do not fall under any of the above categories.

\subsection{Data annotation}

Since annotating amodal layout from a single image is a hard task even for professional annotators, we leverage multiple sources of information to guide this procedure. We first build a point cloud map of the entire scene, annotate the map (in full 3D), and then generate the other views (\bev{}).
To build a map of the entire area, we use the lidar scans corresponding to each frame and register them to a global view. We use the precise odometry information provided in KITTI for registration, to be robust to outliers, as many scenes contain dynamic objects (vehicles/pedestrians).
To reduce manual effort, we semi-automate the data annotation pipeline to the extent possible. As a first step, we identify lane markers by choosing a range of remission values of the lidar and thresholding the 3D map.
Once a set of lane markers are identified, we tag as \emph{lanes} the area between pairs of lane markers. Manual verification is performed to ensure no discrepancies creep in.
Each identified \emph{lane} is labeled with a different \emph{lane id}, unique within a given sequence.
Annotators are then presented with this preprocessed map in an orthographic view and asked to amodally complete annotations. Specifically, chunks of missing points in a lane are \emph{filled in} as \emph{lanes}, and so on. Other identified chunks are annotated as \emph{sidewalk}, \emph{crosswalk}, or \emph{other road}. We repurpose the annotation tool from Semantic KITTI~\cite{semantickitti} for this.

Note that the above annotation procedure can only be used for static parts of the scene. For dynamic scene components, such as vehicles, a 3D bounding box annotation is performed per-frame. This entire annotated sequence is then projected into \bev{} and semantics are made available in every camera frame in the sequence.

 \section{Approach}
\label{sec:approach}

\begin{figure*}[!ht]
  \centering
  \includegraphics[width=\textwidth]{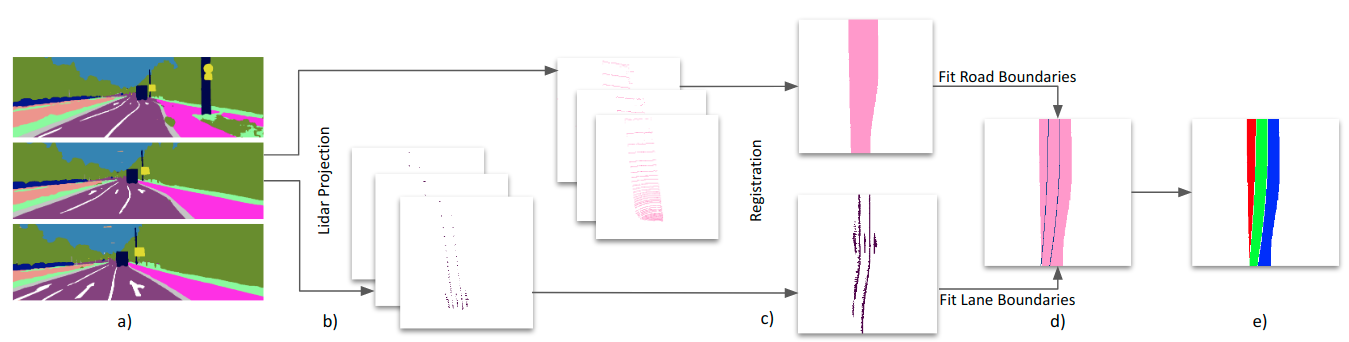}
  \caption{\textbf{Weak Supervision setup}: We show our automated (noisy) label generation scheme herein. Points from the lidar frame are projected down to semantically segmented images (a), to obtain sparse static layouts (b). These sparse layouts are stacked across an image sequence to generate dense static layouts (c). In the next step (d), road and lane boundaries are extracted and combined to obtain lane layouts (e).}
  \label{fig: weak_supervision}
\end{figure*}





Learning the mapping function described in Sec.~\ref{sec:problem} which maps the perspective RGB images to the static and dynamic layouts in \bev{} is quite a challenging problem. The problem demands an accurate understanding and modeling of the 3D scene which necessitates the requirement for learning good visual features that can encode the depth and the semantics of the 3D scene. Most approaches~\cite{schulter2018learning, monolayout, monooccupancy} to monocular layout estimation operate on a single RGB image input, which often leads to temporally inconsistent predictions.

In this section, we tackle the aforementioned problem by proposing a neural network architecture that takes a sequence of images as input and generates temporally consistent layouts as output. 



\subsection{\videolayout{}: Temporally Consistent Layout Estimation}
\label{subsec:videolayout}


Given a sequence of RGB images $\mathcal{I}_1, \mathcal{I}_2, ..., \mathcal{I}_t$ as input, our model predicts the posterior distribution $P(\mathcal{S}_t, \mathcal{D}_t | {\mathcal{I}_1, \mathcal{I}_2, ..., \mathcal{I}_t})$, where $\mathcal{S}_t$ and $\mathcal{D}_t$ denote the predicted static and dynamic layouts corresponding to image $\mathcal{I}_t$.

Our network architecture consists of four subparts:
\begin{itemize}
    \item A feature encoder which takes as input a sequence of RGB images $\mathcal{I}_1, \mathcal{I}_2, ..., \mathcal{I}_t$ and generates rich multi-scale features maps $\mathcal{C}_1, \mathcal{C}_2, ..., \mathcal{C}_t$ corresponding to each rgb image.
    
    \item A stacked Convolutional LSTM submodule which aggregates image features $\mathcal{C}_1, \mathcal{C}_2, ..., \mathcal{C}_t$  and encodes a temporal representation useful for estimating consistent layouts. 
    
    \item Similar to \cite{monolayout}, we maintain two set of decoder weights corresponding to the static and dynamic decoders. Both decoders share a similar architecture except for the use of sampling layer at the beginning of the static decoder in order to generate smooth looking road and lane layouts. Instead of directly processing the encoded output representation of convLSTM, we instead sample from a gaussian distribution with the feature map as the mean values and a fixed standard deviation. 
    
    
    \item A Refinement Network, which helps in improving the quality of the output static and dynamic layouts by regularizing the predicted layouts in order to resemble the true data distribution.
\end{itemize}



\subsection{Consistency Loss}

The weights of the \videolayout{} architecture are updated via backpropagation using the loss function defined in Eq.~\ref{eq:total_loss}. 

\vspace{-1em}
\begin{equation}
    \mathcal{L}_{sup} = \sum_{i=1}^N f(\Bar{S}_i,  S_i) +  f(\Bar{D}_i,  D_i) 
\end{equation}

\vspace{-1.5em}
\begin{equation}
    \mathcal{L}_{short}^{c} = \sum_{i=1}^N \sum_{j=1}^{seqlen-1} f(\Bar{S}_i^j, \Bar{S}_i^{j+1}) + f(\Bar{D}_i^j, \Bar{D}_i^{j+1})
\end{equation}

\vspace{-1.5em}
\begin{equation}
    \mathcal{L}_{long}^{c} = \sum_{i=1}^N \sum_{j=1}^{seqlen-1} \sum_{k=j+2}^{seqlen} f(\Bar{S}_i^j, \Bar{S}_i^{k}) + f(\Bar{D}_i^j, \Bar{D}_i^{k})
\end{equation}

\vspace{-1.5em}
\begin{equation}
    \mathcal{L} = \lambda_{sup} * \mathcal{L}_{sup} + \lambda_{short}^c * \mathcal{L}_{short}^{c} + \lambda_{long}^c * \mathcal{L}_{long}^{c}
    \label{eq:total_loss}
\end{equation}

Here, $\Bar{S}$ and $\Bar{D}$ denote the predicted static and dynamic layouts respectively, $S$ and $D$ denote the ground truth (weak/strong) layouts. $N$ is the mini batch size, $seqlen$ is the length of the input image sequence per sample. $f(\cdot, \cdot)$ is the cross-entropy loss function. $\mathcal{L}_{sup}$ is a supervised loss term that penalizes the deviation of the predicted static and dynamic layouts. $\mathcal{L}_{short}^{c}$ is the short-range consistency loss and  $\mathcal{L}_{long}^{c}$ is the long-range consistency loss. $\lambda_{sup}$, $\lambda_{short}^c$ and $\lambda_{long}^c$ are the weights corresponding to the supervised, short-range consistency and long-range consistency losses respectively. Finally, $\mathcal{L}$ is the total weighted loss used for backpropagating gradients through the network. \footnote{$\lambda_{sup} > \lambda_{short}^{c} >> \lambda_{long}^c$}. Consistency loss acts as a soft constraint on the layout prediction objective $\mathcal{L}_{sup}$, which enables the model to learn more temporally consistent layout.



\subsection{Refinement Network}


Similar to \cite{monolayout}, we use adversarial regularization in order to restrict the encoder-decoder network into predicting conceivable layouts. For adversarial regularization, we use a Patch Discriminator which takes the output of the encoder-decoder and a sample from the true data distribution. Unlike \cite{monolayout}, we make use of ground-truth layouts to do paired training with the discriminator which further improves the performance of the model. Adding the sampling layer in the static decoder converts the Encoder-Decoder network into a generator and further helps in regularization for static layouts.

\subsection{Data Preparation: Weak Supervision}
\label{seubsec:dataprep}

Since the amodal scene layout labels are not provided for almost every autonomous driving dataset, we propose a generic layout generation method to generate noisy or weak supervision layout labels for static classes like roads, lanes, and sidewalks (\cf{}~Fig.~\ref{fig: weak_supervision}). Our weak supervision layout generation method is similar to the sensor fusion approach adopted in \cite{monolayout}. Further, we provide a solution to obtain weak supervision layout labels for lanes as well. 

The process of aggregating / registering the lidar points for several image frames and projecting them to an orthographic view by assuming a flat plane is shown in Fig.~\ref{fig: weak_supervision}. To obtain lanes, we perform the registering process for road and lane marker semantic classes. We then obtain lane boundary curves by isolating individual lane boundaries through Density Based Spatial Clustering and fit third order polynomial curves on these individual lane boundaries. From the registered road layout, we estimate the road boundaries and group the road and lane boundaries together to obtain lane contours.

Additionally, we also obtain improved weak supervision layout for roads by addressing cases where there are too many stationary obstacles in a sequence occluding lidar from capturing the road layout. We register lidar points belonging to those stationary obstacles that lie on the road.

\section{Experiments}

\begin{figure*}[!ht]
  \centering
    \begin{adjustbox}{max width=0.9\linewidth, totalheight=13cm}  \includegraphics[width=0.8\linewidth]{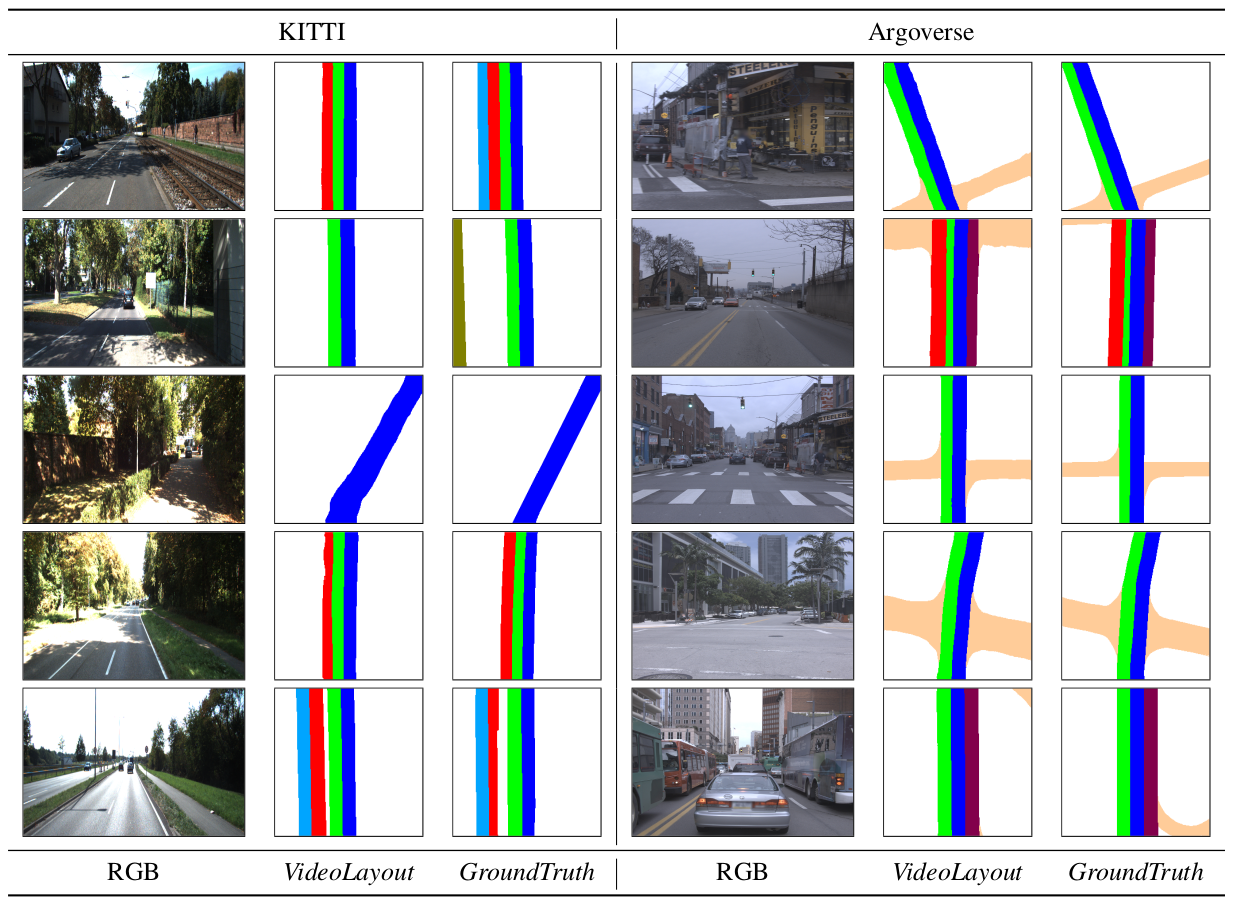}
    \end{adjustbox}
    \caption{\footnotesize{\textbf{Static layout estimation}: \videolayout{} predicts fine-grained attributes of the road scene including lanes, side roads and ego-lane. Each individual color represents a single lane. Ego-lane is shown in {\textcolor{blue}{blue}} and side-roads are shown in  light  {\textcolor{orange}{orange}}. \videolayout{} produces decent static layout estimates for both AutoLay and Argoverse datasets. Its able to hallucinate occluded regions in the scene very reliably specially for Argoverse dataset.}}
  \label{fig:qualitative_road}
  \vspace{-0.5cm}
\end{figure*}


So far, there has been a lack of consensus on datasets and evaluation protocol for amodal layout estimation. We describe the \coolname{} evaluation protocol in detail, and evaluate baselines and existing approaches on the benchmark.

\subsection{Datasets}


To ensure fair comparison with existing methods, we show our results on the Argoverse~\cite{chang2019argoverse} and KITTI RAW split used in~\cite{monolayout} and~\cite{schulter2018learning} respectively. Argoverse~\cite{chang2019argoverse} provides detailed high-resolution semantic maps in \bev{} for roads, vehicles and lanes, which makes it suitable for evaluation on all the tasks proposed in Sec.~\ref{sec:problem}. We present a breakdown of the \coolname{} benchmark below.

\begin{enumerate}
    \item \textbf{\emph{KITTI} split}: The \emph{KITTI} split of \autolay{} comprises $10705$ train images and $5626$ test images. We report results for models trained using weak (proposed data preparation) and strong (ground-truth) supervision.
    \item \textbf{\emph{KITTI RAW} split}: For fair evaluation with Schulter~\etal{}~\cite{schulter2018learning}, we use the KITTI RAW split comprising $16269$ train images and $2305$ test images).
    \item \textbf{\emph{Argoverse} split}: The Argoverse~\cite{chang2019argoverse} split contains $6722$ train images and $2418$ test images.
\end{enumerate}

\subsection{Implementation Details}

\subsubsection{PseudoLidar input}
We evaluate the performance of a set of approaches that take PseudoLidar\cite{wang2019pseudo} as input for amodal scene layout estimation. The PseudoLidar representation is a processed point cloud obtained by projecting the RGB image pixels to the camera coordinate system using the per-pixel depth values obtained from unsupervised monocular depth estimation methods like \cite{godard2018digging}. We substitute the remission value by the normalized mean of RGB channel intensities. We then obtain a voxel representation of the point cloud by slicing it vertically along the X-Z plane with a resolution of 0.15625 m and horizontally along the Y-axis into 10 channels starting from 0.4 m above the camera to 2 m below the camera. This PseudoLidar representation is provided as the input to ENet \cite{enet} or Unet \cite{unet} for amodal scene layout estimation.

\subsubsection{Lane layout convention}
In order to convert lane estimation, which is essentially an instance segmentation/detection task in \bev{} to an occupancy prediction task, we adhere to the following convention while performing lane layout estimation. While creating the per frame training labels using both weak and strong supervision, we classify the individual lane layouts on the road with the ego-vehicle, (\emph{ego-road}), as either  \emph{ego lane} or \emph{other lanes}. The ego lane and the other lanes are provided individual lane ids. The lane ids to the other lanes are given on the basis of their position with respect to the ego-lane.


\subsection{Evaluation Metrics}

\subsubsection{Vehicle Layout}
Similar to \cite{monolayout}, We evaluate the vehicle layouts on both mean intersection-over-union(mIoU) and mean average-precision(mAP). Since we are evaluating different methods on the task of vehicle occupancy prediction in \bev{}, we don't go for the traditional evaluation metric($AP_{IoU>0.7}$ or $AP_{IoU>0.5}$) used by 3D object detection methods or object detection methods in \bev{}.

\subsubsection{Road Layout}

We adopt mean intersection-over-union(mIoU) and mean average-precision(mAP) as our metric for evaluating road layouts on the datasets. To evaluate hallucination capabilities of different methods, we also make use of the occluded mean intersection-over-union (IoU) used in \cite{schulter2018learning}, \cite{monolayout}. This involves calculating IoU for only the portions of the road which are occluded. 

\subsubsection{Lane Layout}
As mentioned in Sec.~\ref{sec:problem}, Lane Layouts are evaluated on two separate tasks: Ego-lane estimation and Overall lane detection. 

\begin{enumerate}
    \item \textit{Ego-Lane Estimation:} Ego-Lane estimation task is evaluated on the mean intersection-over-union (mIoU) and mean average-precision (mAP) metric.
    
    \item \textit{Overall Lane Detection:} For this task, we use $AP_{IoU>0.7}$, a popular metric used for evaluating object detection and instance segmentation methods. A predicted lane is counted as a detection if its intersection-over-union(IoU) with one of the ground truth lanes is greater than 0.7(70\%). $AP_{IoU>0.7}$ provides information about the average-precision of predicted detections. We also report $Recall_{IoU>0.7}$, which gives us an idea about the prediction capability of the method. 
\end{enumerate}

\subsection{Evaluated Methods}
We evaluate the performance of the following approaches.
\begin{itemize}
    \item \emph{Schulter~\etal{}}: The static scene layout estimation approach proposed in~\cite{schulter2018learning}.
    \item \emph{MonoOccupancy}: The static scene layout estimation approach proposed in~\cite{monooccupancy}.
    \item \emph{MonoOccupancy-ext}: We extend MonoOccupancy~\cite{monooccupancy} to predict vehicle occupancies.
    \item \emph{PseudoLidar-ENet}: A ENet~\cite{enet} architecture with PseudoLidar input for amodal scene layout estimation.
    \item \emph{PseudoLidar-UNet}: A UNet~\cite{unet} architecture with PseudoLidar input for amodal scene layout estimation.
    \item \emph{MonoLayout}: Amodal scene layout estimation architecture proposed in~\cite{monolayout}
    \item \videolayout: The full \videolayout{} architecture trainned with temporal-consistency loss and the adversarial refinement network.
\end{itemize}

\label{sec:experiments}
\begin{table*}[!ht]
\begin{center}
\begin{adjustbox}{max width=\linewidth}
\begin{tabular}{c|c|c|c|c|c|c|c|c|c}
\centering

                                                     &                                              & \multicolumn{2}{c|}{\textbf{Vehicle Layout}} & \multicolumn{6}{c}{\textbf{Static Layout Estimation}}       \\ 
                                                    \hline
                                                     &                                              &           &                   & \multicolumn{2}{c|}{\textbf{Road}}   & \multicolumn{4}{c}{\textbf{Lane}}  \\ 
                                                    \cline{5-10} 
                                    \textbf{Dataset}          &          \textbf{Method}                              &           &                   &             &             &  \multicolumn{2}{c|}{\textbf{Ego-Lane}}    &      \multicolumn{2}{c}{\textbf{Overall}}   \\
                                                     \cline{7-10}
                                                     &                                              & \textbf{mIoU}      &     \textbf{mAP}           &  \textbf{mIoU}       & \textbf{mAP}         &   \textbf{mIoU}    &   \textbf{mAP}   &      \textbf{\small{$AP_{\small{iou>0.7}}$}}   &   \textbf{\small{$Recall_{iou>0.7}$}}   \\
                                                    \cline{1-10} 
\multirow{5}{*}{KITTI} & PseudoLidar-UNet  & $20.10$   &      $52.47$      & $57.62$     &  $70.90$    &   $51.98$  &   $68.94$   &  $8.53$ &  $19.09$         \\
                         & PseudoLidar-ENet  & $15.26$   &      $40.05$      & $55.70$     &  $67.62$    &   $54.42$  &   $68.15$   &   $13.01$   &  $24.50$     \\
                         & MonoOccupancy-ext  & $27.03$   &      $42.92$      & $63.10$     &  $77.56$    &  $58.39$  &  $69.16$   &   $26.11$   &  $35.36$         \\
             & MonoLayout\cite{monolayout}   & $30.88$   &      $53.63$      & $63.83$     &  $77.53$    &   $58.27$  &  $72.08$   &   $16.09$  &  $34.91$     \\                                                     
                        & VideoLayout        & $\mb{34.60}$   & $\mb{52.87}$      & $\mb{68.73}$     &  $\mb{84.89}$    &   $\mb{62.68}$  &  $\mb{76.80}$   &     $\mb{38.56 }$  &  $\mb{44.33}$   \\                                                     
                                                    \hline
\multirow{3}{*}{Argoverse} & MonoOccupancy-ext   & $16.22$   &      $38.66$      &  $72.41$     &  $79.62$    &   $75.26$   & $85.51$   &     $39.76$   &  $51.65$     \\
                & MonoLayout\cite{monolayout}    & $28.31$   &      $46.07$      & $73.25$     &  $84.56$    &   $71.73$   & $82.00$   &     $26.18$   &  $42.61$       \\                       
                     & VideoLayout               & $\mb{32.77}$   &      $\mb{50.48}$      & $\mb{76.42}$     &  $\mb{88.01}$    &   $\mb{77.62}$   & $\mb{87.77}$   &     $\mb{46.03}$   &  $\mb{54.86}$      \\      
\end{tabular}
\end{adjustbox} 
\end{center}
\caption{\textbf{Quantitative results}:
We benchmark \videolayout{}, \monolayout{} \cite{monolayout}, MonoOccupancy-ext and PseudoLidar based approaches on KITTI and Argoverse datasets.}
\label{table:quantitative:main}
\end{table*}


\begin{table}[]
    \centering
    \begin{adjustbox}{max width=\textwidth, totalheight=2.5cm}
    \begin{tabular}{|c|c|c|c|c|}
        \hline
                                                      &             &                  &  Road     &         \\

                                                      & Road        & Sidewalk         &   +      &  Lane       \\
            Method                                    &             &                  & Sidewalk &              \\
        \cline{2-5}
                                                      & mIOU        &  mIOU            &    occ mIOU     &  \small{$AP_{\small{iou>0.7}}$}  \\
        \hline
        MonoOccupancy\cite{monooccupancy}             & $56.16$       &  $18.18$       &   $28.24$       &   $26.64$          \\
        Schulter \etal{} \cite{schulter2018learning}  & $68.89$       &  $30.35$       &   $61.06$       &      -          \\
        MonoLayout\cite{monolayout}                   & $73.86$       &  $32.86$       &   $67.42$      &    $16.20$       \\
        \videolayout{}                                & $\mb{75.92}$  &  $\mb{37.74}$  &   $\mb{71.01}$  &  $\mb{36.64}$                 \\
        \hline
    \end{tabular}
    \end{adjustbox}
    \caption{State-of-the-art comparision for static layout(Road, Sidewalk and Lane) with weak supervision(\cf{} Sec. \ref{seubsec:dataprep}) on KITTI RAW split used in \cite{schulter2018learning} }
    \label{table:kittiraw}
\end{table}

\begin{figure}
    \centering
    \includegraphics[width=\linewidth]{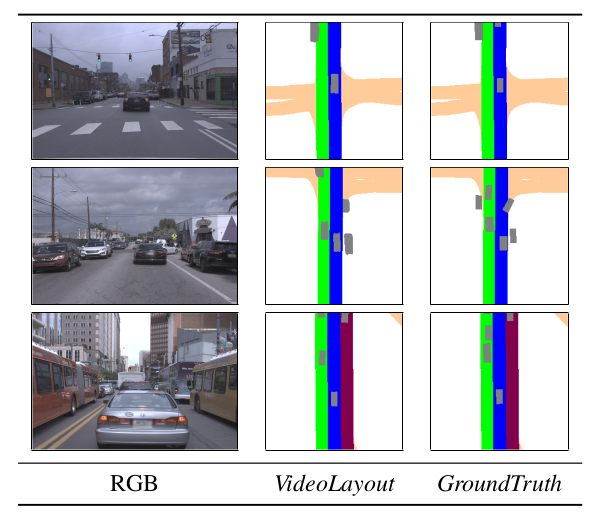}
    \caption{\textbf{Dynamic Layout Estimation:}. \videolayout{} provides crisp and accurate vehicle occupancies. The grey boxes indicate vehicle occupancies. Observe the ability of \videolayout{} to precisely localize vehicles that are distant and partially occluded. }
    \label{fig:dynamic}
\end{figure}

\subsection{Road Layout Estimation}
\label{subsec:roadlayoutest}

We evaluate PseudoLidar based (PseudoLidar-UNet, PseudoLidar-ENet) and RGB image based architectures (MonoOccupancy \cite{monooccupancy}, \monolayout{} and \videolayout{}) for the task of road layout estimation using the ground truth labels. Table~\ref{table:quantitative:main} provides a detailed benchmark of these methods on the \autolay{} and Argoverse\cite{chang2019argoverse} datasets. We also perform state-of-the-art comparision(Table~\ref{table:kittiraw}) with previously proposed methods in literature on KITTI RAW split proposed in  \cite{schulter2018learning} on the task of road and lane layout estimation. For this comparision, we use the weak data generation method mentioned in Sec~\ref{seubsec:dataprep}.  

We observe that \monolayout{} and \videolayout{} perform better than PseudoLidar-based methods for road layout estimation. This can be attributed to the sparsity of the PseudoLidar points as the distance from camera increases. Also, PseudoLidar based inputs don't allow the use of deep feature encoders like ResNet\cite{he2016deep} needed to extract the rich visual features necessary for hallucinating amodal scene layouts. From Table~\ref{table:quantitative:main} it can be seen that MonoOccupancy-ext performs better than the PseudoLidar counterparts.
This further exemplifies the narrative that with better supervision, an RGB image-based architecture can reason for occluded regions better than pseudo-lidar based approaches.

Within the RGB image based approaches, we observe that \videolayout{} outperforms \monolayout{}, MonoOccupancy \cite{monooccupancy}  by considerable margins for all the evaluation metrics. The same can be also be observed in the results on KITTI RAW (\cf{}~Table~\ref{table:kittiraw}). 


\subsection{Vehicle Layout Estimation}

From Table~\ref{table:quantitative:main}, we can see that PseudoLidar-based methods have inferior performance in comparision to monocular methods\cite{monolayout, monooccupancy} on vehicle layout estimation. This can again be attributed to the fact that monocular depths are inprecise and sparse at large distances, which makes the PseudoLidar input unreliable for this task. 

Similar to road layout estimation, \videolayout{} performs better than \monolayout{} due to the temporal consistency loss and the adversarial regularization via the refinement network (\cf{} Fig.~\ref{fig:dynamic}). Also, MonoOccupany~\cite{monooccupancy} performs poorly on vehicle layout estimation, presumably due to \emph{blurriness} in the variational autoencoder used therein leading to low mAP scores.

\subsection{Lane Layout Estimation}

We split the lane layout estimation into two tasks: \emph{ego-lane estimation} and \emph{overall lane detection}. This exclusive treatment of ego-lane estimation is due to its prominence in semi and fully autonomous driving.
Unlike objects, which can be ``detected" as bounding boxes, lanes are objects whose extents span potentially the entire width/height of the \bev{} image, and this calls for an appropriate evaluation protocol. We thus treat lane estimation akin to a ``segmentation" problem, and compare performance in terms of average precision and recall at varying IoU thresholds ($0.5$, $0.7$).
From Table~\ref{table:quantitative:main}, we see that that MonoOccupancy~\cite{monooccupancy} performs significantly better than PseudoLidar methods on ego-lane estimation.

\videolayout{} significantly outperforms other methods, owing to its ability to encode temporal information and the use of a separate refinement network. The refinement network corrects blobby lane estimates that do not match the true data distribution, resulting in performance boosts at higher IoU thresholds. A few qualitative results are presented in Fig.~\ref{fig:qualitative_road}.


\begin{table}[]
    \centering
    \begin{adjustbox}{max width=\linewidth}
    \begin{tabular}{|c|c|c|c|c|}
        \hline  
                                      & \multicolumn{2}{c|}{Ego-Lane}     &  \multicolumn{2}{c|}{OverAll}       \\
        \cline{2-5}
        Method                        & mIOU        &  mAP               &  \small{$AP_{\small{iou>0.7}}$}    & \small{$R_{\small{iou>0.7}}$}   \\
        \hline
        VideoLayout (no samp., no reg.)   & $59.81$       &  $73.13$         &   $20.95$     &    $36.55$  \\
        VideoLayout (no reg.)             & $61.15$       &  $75.72$         &   $35.89$     &    $41.92$    \\
        VideoLayout (full)                & $62.68$       &  $76.80$         &   $38.56$     &    $44.33$           \\ 
        \hline
    \end{tabular}
    \end{adjustbox}
    \caption{Analyzing effect of various subparts of the \videolayout{} architecture on Lane Layout estimation performance. The sampling layer and adversarial regularization improve performance.}
    \label{table:sampling} 
\end{table}

\begin{table}[]
    \centering
    \begin{adjustbox}{max width=\textwidth, totalheight=1.5cm}
    \begin{tabular}{|c|c|c|c|c|}
        \hline
        Method              &  {\tiny \small{$AP_{\small{iou>0.5}}$}}  &   \small{$R_{\small{iou>0.5}}$} &  \small{$AP_{\small{iou>0.7}}$}   & \small{$R_{\small{iou>0.7}}$}   \\
        \hline
        PseudoLidar-UNet             & $19.32$       &  $43.24$         &   $8.53$     &     $19.09$  \\
        PseudoLidar-ENet             & $25.10$       &  $47.39$         &   $13.01$     &    $24.50$    \\
        MonoLayout\cite{monolayout}  & $25.38$       &  $55.06$         &   $16.09$     &    $34.91$           \\
        VideoLayout                  & $57.02$       &  $65.49$         &   $38.56$     &    $44.33$   \\
        \hline
    \end{tabular}
    \end{adjustbox}
    \caption{Analyzing Lane layout scores at IoU thresholds of $0.5$ and $0.7$. $AP$ denotes average precision. $R$ denotes Recall. \videolayout{} consistently performs better across IoU thresholds.}
    \label{table:iou_thresh}
\end{table}

\subsection{Ablation Studies}

We conduct ablation studies to analyze the performance of various components in the pipeline. These lead to a number of interesting observations that we enlist below.

\subsubsection{Sensitivity of lane estimates to IoU thresholds}

In Table~\ref{table:iou_thresh}, we analyze the sensitivity of lane estimates to IoU thresholds. At lower thresholds ($AP_{IoU>0.5}$ and $R_{IoU>0.5}$), PseudoLidar-based techniques have similar performance to that of MonoLayout~\cite{monolayout}.
But as the threshold increases ($AP_{IoU>0.7}$ and $R_{IoU>0.7}$), PseudoLidar methods exhibit a dramatic performance drop, particularly in terms of recall.
We attribute this to the fact that PseudoLidar relies on monocular depth estimates which are sparse and inaccurate for vehicles farther from the camera.
On the other hand, \videolayout{} performs consistently across both IoU thresholds. Particularly at $AP_{IoU>0.5}$, we observe a significant performance increase compared to \monolayout{}.


\subsubsection{Effect of Sampling and Adversarial Refinement}

Table~\ref{table:sampling} shows the performance of different variants of \videolayout{} for lane layout estimation. \videolayout{} (no samp., no reg.) refers to a variant without a sampling layer in the static decoder and without adversarial regularization. \videolayout{} (no reg.) uses sampling layer in the static decoder. \videolayout{} (full) uses both sampling layer and an adversarial regularizer. It can be seen that each component meaningfully contributes to performance boosts.
The sampling layer enhances the capability of the model to capture the distribution of \emph{plausible} layouts, improving precision ($AP_{IoU>0.7}$).
The adversarial regularizer improves the sharpness of the predicted samples, improving the recall $R_{IoU>0.7}$.

\begin{figure}[!t]
    \centering
    \begin{adjustbox}{max width=\linewidth}
    \includegraphics[]{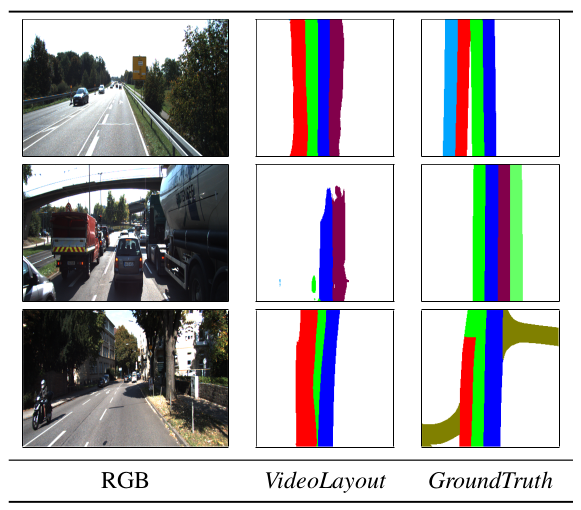}
    \end{adjustbox}
    \caption{\footnotesize{\textbf{Failure Cases} of \videolayout{}. (\emph{Top row}) Failure to predict ego-lane. (\emph{Middle row}) Failure in heavy traffic scenarios. (\emph{Bottom row}) Failure in predicting lane width. }}
    \label{fig:failure}
\end{figure}

\subsection{Failure cases}

Despite its impressive performance, \videolayout{} fails in multiple scenarios. A few representative failure cases are shown in Fig.~\ref{fig:failure}. Often, such failures are due to high-dynamic range conditions, where shadows are mistaken for lanes, or in heavy traffic scenarios.

\subsection{Runtime}

On an NVIDIA RTX 1080Ti, the inference rate of \videolayout{} is slightly over $40$ frames per second, which is suitable for real-time performance.

\section{Conclusion}

Amodal layout estimation is an emerging perception task for autonomous driving. Solving this task necessitates an understanding of not just entities present in an image, but also a reasoning of occluded portions. Benchmarks have driven progress in allied perception tasks such as object detection, segmentation, tracking, depth estimation, and more. In that spirit, we introduce a new benchmark (\coolname{}) for amodal layout estimation. We implement several baselines and prior art, and make all our code and data available. We further propose \videolayout{}, a simple yet effective technique, and demonstrate that reasoning about image sequences leads to coherent and robust layout estimates. We hope the dataset and benchmark serve as a breeding ground for a new class of approaches that will take us a step closer to understanding the 3D world from a sequence of its projections.




\bibliography{main}

\begin{thebibliography}{10}
\providecommand{\url}[1]{#1}
\csname url@rmstyle\endcsname
\providecommand{\newblock}{\relax}
\providecommand{\bibinfo}[2]{#2}
\providecommand\BIBentrySTDinterwordspacing{\spaceskip=0pt\relax}
\providecommand\BIBentryALTinterwordstretchfactor{4}
\providecommand\BIBentryALTinterwordspacing{\spaceskip=\fontdimen2\font plus
\BIBentryALTinterwordstretchfactor\fontdimen3\font minus
  \fontdimen4\font\relax}
\providecommand\BIBforeignlanguage[2]{{%
\expandafter\ifx\csname l@#1\endcsname\relax
\typeout{** WARNING: IEEEtran.bst: No hyphenation pattern has been}%
\typeout{** loaded for the language `#1'. Using the pattern for}%
\typeout{** the default language instead.}%
\else
\language=\csname l@#1\endcsname
\fi
#2}}

\bibitem{KITTI}
A.~Geiger, P.~Lenz, and R.~Urtasun, ``Are we ready for autonomous driving? the
  kitti vision benchmark suite,'' in \emph{CVPR}, 2012.

\bibitem{chang2019argoverse}
M.-F. Chang, J.~Lambert, \emph{et~al.}, ``Argoverse: 3d tracking and
  forecasting with rich maps,'' in \emph{CVPR}, 2019.

\bibitem{monolayout}
K.~Mani, S.~Daga, \emph{et~al.}, ``Monolayout: Amodal layout estimation from a
  single image,'' \emph{IEEE Winter Conference on Applications of Computer
  Vision (WACV)}, 2020.

\bibitem{monooccupancy}
C.~Lu, M.~J.~G. van~de Molengraft, and G.~Dubbelman, ``Monocular semantic
  occupancy grid mapping with convolutional variational encoder-decoder
  networks,'' \emph{IEEE Robotics and Automation Letters}, 2019.

\bibitem{schulter2018learning}
S.~Schulter, M.~Zhai, N.~Jacobs, and M.~Chandraker, ``Learning to look around
  objects for top-view representations of outdoor scenes,'' in \emph{ECCV},
  2018.

\bibitem{you2019pseudo}
Y.~You, Y.~Wang, \emph{et~al.}, ``Pseudo-lidar++: Accurate depth for 3d object
  detection in autonomous driving,'' \emph{arXiv preprint}, 2019.

\bibitem{ozyecsil2017survey}
O.~{\"O}zye{\c{s}}il, V.~Voroninski, R.~Basri, and A.~Singer, ``A survey of
  structure from motion*.'' \emph{Acta Numerica}, vol.~26, pp. 305--364, 2017.

\bibitem{amodalKTCM15}
A.~Kar, S.~Tulsiani, J.~Carreira, and J.~Malik, ``Amodal completion and size
  constancy in natural scenes,'' in \emph{International Conference on Computer
  Vision (ICCV)}, 2015.

\bibitem{factored3dTulsiani17}
S.~Tulsiani, S.~Gupta, \emph{et~al.}, ``Factoring shape, pose, and layout from
  the 2d image of a 3d scene,'' in \emph{Computer Vision and Pattern
  Regognition (CVPR)}, 2018.

\bibitem{wang2019parametric}
Z.~Wang, B.~Liu, S.~Schulter, and M.~Chandraker, ``A parametric top-view
  representation of complex road scenes,'' in \emph{CVPR}, 2019.

\bibitem{lu2019hallucinating}
C.~Lu and G.~Dubbelman, ``Hallucinating beyond observation: Learning to
  complete with partial observation and unpaired prior knowledge,'' \emph{arXiv
  preprint}, 2019.

\bibitem{garnett20193d}
N.~Garnett, R.~Cohen, \emph{et~al.}, ``3d-lanenet: end-to-end 3d multiple lane
  detection,'' in \emph{Proceedings of the IEEE International Conference on
  Computer Vision}, 2019, pp. 2921--2930.

\bibitem{ku2018joint}
J.~Ku, M.~Mozifian, \emph{et~al.}, ``Joint 3d proposal generation and object
  detection from view aggregation,'' in \emph{IROS}, 2018.

\bibitem{chen2017multi}
X.~Chen, H.~Ma, \emph{et~al.}, ``Multi-view 3d object detection network for
  autonomous driving,'' in \emph{CVPR}, 2017.

\bibitem{liang2018deep}
M.~Liang, B.~Yang, S.~Wang, and R.~Urtasun, ``Deep continuous fusion for
  multi-sensor 3d object detection,'' in \emph{ECCV}, 2018.

\bibitem{yang2018pixor}
B.~Yang, W.~Luo, and R.~Urtasun, ``Pixor: Real-time 3d object detection from
  point clouds,'' in \emph{CVPR}, 2018.

\bibitem{roddick2018orthographic}
T.~Roddick, A.~Kendall, and R.~Cipolla, ``Orthographic feature transform for
  monocular 3d object detection,'' \emph{arXiv preprint}, 2018.

\bibitem{wang2019pseudo}
Y.~Wang, W.-L. Chao, \emph{et~al.}, ``Pseudo-lidar from visual depth
  estimation: Bridging the gap in 3d object detection for autonomous driving,''
  in \emph{CVPR}, 2019.

\bibitem{oxfordrobotcar}
W.~Maddern, G.~Pascoe, C.~Linegar, and P.~Newman, ``{1 Year, 1000km: The Oxford
  RobotCar Dataset},'' \emph{The International Journal of Robotics Research
  (IJRR)}, vol.~36, no.~1, pp. 3--15, 2017.

\bibitem{nuscenes2019}
H.~Caesar, V.~Bankiti, \emph{et~al.}, ``nuscenes: A multimodal dataset for
  autonomous driving,'' \emph{arXiv preprint arXiv:1903.11027}, 2019.

\bibitem{apolloscape}
X.~Huang, P.~Wang, \emph{et~al.}, ``The apolloscape open dataset for autonomous
  driving and its application,'' \emph{arXiv preprint arXiv:1803.06184}, 2018.

\bibitem{OpenStreetMap}
{OpenStreetMap contributors}, ``{Planet dump retrieved from
  https://planet.osm.org },'' \url{ https://www.openstreetmap.org }, 2017.

\bibitem{semantickitti}
J.~Behley, M.~Garbade, \emph{et~al.}, ``Semantickitti: A dataset for semantic
  scene understanding of lidar sequences,'' in \emph{ICCV}, 2019.

\bibitem{godard2018digging}
C.~Godard, O.~Mac~Aodha, M.~Firman, and G.~Brostow, ``Digging into
  self-supervised monocular depth estimation,'' \emph{arXiv preprint}, 2018.

\bibitem{enet}
A.~Paszke, A.~Chaurasia, S.~Kim, and E.~Culurciello, ``Enet: A deep neural
  network architecture for real-time semantic segmentation,'' \emph{arXiv
  preprint arXiv:1606.02147}, 2016.

\bibitem{unet}
O.~Ronneberger, P.~Fischer, and T.~Brox, ``U-net: Convolutional networks for
  biomedical image segmentation,'' in \emph{International Conference on Medical
  image computing and computer-assisted intervention}, 2015.

\bibitem{he2016deep}
K.~He, X.~Zhang, S.~Ren, and J.~Sun, ``Deep residual learning for image
  recognition,'' in \emph{Proceedings of the IEEE conference on computer vision
  and pattern recognition}, 2016, pp. 770--778.

\end{thebibliography}
\bibliographystyle{styles/IEEEtran}

\end{document}